\documentclass[sigconf]{acmart}

\AtBeginDocument{%
  \providecommand\BibTeX{{%
    \normalfont B\kern-0.5em{\scshape i\kern-0.25em b}\kern-0.8em\TeX}}}

\usepackage{amsmath,amsfonts,bm}

\def\eqref#1{equation~\ref{#1}}

\def\1{\bm{1}}

\DeclareMathAlphabet{\mathsfit}{\encodingdefault}{\sfdefault}{m}{sl}
\SetMathAlphabet{\mathsfit}{bold}{\encodingdefault}{\sfdefault}{bx}{n}

\def\sA{{\mathbb{A}}}

\def\sD{{\mathbb{D}}}

\def\sS{{\mathbb{S}}}

\newcommand{\E}{\mathbb{E}}

\DeclareMathOperator*{\argmax}{arg\,max}

\usepackage{amsmath}  
\usepackage{amsthm}   
\usepackage{amsfonts} 
\usepackage{dsfont}   

\usepackage{algorithm} 
\usepackage{algorithmic} 
\renewcommand{\algorithmiccomment}[1]{\bgroup\hfill//~#1\egroup} 

\usepackage{pifont} 
\usepackage{array} 
\newcolumntype{L}[1]{>{\raggedright\arraybackslash}m{#1}}  
\newcolumntype{C}[1]{>{\centering\arraybackslash}m{#1}} 
\newcolumntype{R}[1]{>{\raggedleft\arraybackslash}m{#1}}  
\usepackage{tablefootnote} 

\usepackage{graphicx}

\usepackage{subcaption} 

\usepackage{printlen} 

\def\sA{{\mathcal{A}}}

\def\sD{{\mathcal{D}}}

\def\sS{{\mathcal{S}}}

\usepackage{systeme,mathtools}
\usepackage{multirow} 
\usepackage{makecell} 
\usepackage{footnote}

\usepackage{xcolor} 
\usepackage{dsfont}
\newcommand{\probP}{\mathds{P}} 

\usepackage{balance}

\copyrightyear{2023}
\acmYear{2023}
\setcopyright{acmlicensed}\acmConference[SIGIR '23]{Proceedings of the 46th International ACM SIGIR Conference on Research and Development in Information Retrieval}{July 23--27, 2023}{Taipei, Taiwan}
\acmBooktitle{Proceedings of the 46th International ACM SIGIR Conference on Research and Development in Information Retrieval (SIGIR '23), July 23--27, 2023, Taipei, Taiwan}
\acmPrice{15.00}
\acmDOI{10.1145/3539618.3592022}
\acmISBN{978-1-4503-9408-6/23/07}

\begin{document}




\title[Model-free Reinforcement Learning with Stochastic Reward Stabilization for Recommender Systems]{Model-free Reinforcement Learning with Stochastic Reward Stabilization for Recommender Systems }



\author{Tianchi Cai}
\affiliation{
    \institution{Ant Group}
    \country{Hangzhou, China}
}
\authornote{Equal contribution.}
\authornote{Corresponding author. <tianchi.ctc@antgroup.com>}

\author{Shenliao Bao}
\affiliation{
    \institution{Ant Group}
    \country{Hangzhou, China}
}
\authornotemark[1]

\author{Jiyan Jiang}
\affiliation{
    \institution{Tsinghua University}
    \country{Beijing, China}
}
\authornotemark[1]

\author{Shiji Zhou}
\affiliation{
    \institution{Tsinghua University}
    \country{Beijing, China}
}

\author{Wenpeng Zhang}
\affiliation{
    \institution{Ant Group}
    \country{Beijing, China}
}

\author{Lihong Gu}
\affiliation{
    \institution{Ant Group}
    \country{Hangzhou, China}
}

\author{Jinjie Gu}
\affiliation{
    \institution{Ant Group}
    \country{Hangzhou, China}
}

\author{Guannan Zhang}
\affiliation{
    \institution{Ant Group}
    \country{Hangzhou, China}
}

\renewcommand{\shortauthors}{Tianchi Cai et al.}

\begin{abstract}
Model-free RL-based recommender systems have recently received increasing research attention due to their capability to handle partial feedback and long-term rewards. However, most existing research has ignored a critical feature in recommender systems: one user's feedback on the same item at different times is random. The stochastic rewards property essentially differs from that in classic RL scenarios with deterministic rewards, which makes RL-based recommender systems much more challenging. In this paper, we first demonstrate in a simulator environment where using direct stochastic feedback results in a significant drop in performance. Then to handle the stochastic feedback more efficiently, we design two stochastic reward stabilization frameworks that replace the direct stochastic feedback with that learned by a supervised model. Both frameworks are model-agnostic, i.e., they can effectively utilize various supervised models. We demonstrate the superiority of the proposed frameworks over different RL-based recommendation baselines with extensive experiments on a recommendation simulator as well as an industrial-level recommender system.
\end{abstract}


\maketitle

\section{Introduction}

With the increasing volume of information available on the internet, recommender systems are developed to help practitioners discover items of interest by learning and predicting user preferences on different items \cite{cheng2016wide,guo2017deepfm,covington2016deep}. They are successfully applied to various applications, such as news \cite{zheng2018drn}, e-learning \cite{lin2021adaptive}, video \cite{ie2019slateq,chen2019top}, online advertising \cite{zhao2021dear}, marketing \cite{yu2021joint} and e-commerce \cite{zou2019reinforcement,he2020learning}.

Reinforcement learning (RL) has recently gained much research interest in recommender systems \cite{chen2021survey,lin2021adaptive,ie2019slateq,he2020learning}. It learns desired behaviors from interactions with an environment to maximize the long-term cumulative reward \cite{sutton2018reinforcement,afsar2021reinforcement}. Compared to conventional methods such as collaborative filtering \cite{koren2022advances} and deep learning-based methods \cite{cheng2016wide,jannach2017recurrent,zhou2018deep}, RL is capable of handling partial feedback and optimizing long-term experience, hence it is promising in many real-world recommendation scenarios \cite{zheng2018drn,chen2019top,zhao2021dear,cai2023marketing}. 

However, it is very challenging to apply RL to industrial-scale recommender systems serving a tremendous amount of users with diverse preferences concerning a huge item corpus  \cite{chen2019top}. As RL-based recommender systems often treat users as states and items as actions, the state and action spaces are extremely large (typically in millions or billions), making classic RL methods rather sample inefficient \cite{bai2019model,chen2019generative,zou2020pseudo}. In classic RL scenarios, one way to improve sample efficiency is to adopt model-based RL, which directly models the environment dynamics \cite{kaiser2019model,schrittwieser2020mastering}. However, in real-world recommender systems, the transition probability of users is intrinsically hard to estimate \cite{chen2021survey}. In fact, most recognized RL-based recommender systems are based on model-free RL methods
\cite{zhao2021dear,chen2019top,chen2022off}, where auxiliary training techniques are recently introduced to accelerate the learning of representations and improve sample efficiency \cite{liu2020end,chen2021user}. 

\begin{figure*}[tb] 
\begin{center}
\includegraphics[width=0.8\textwidth]{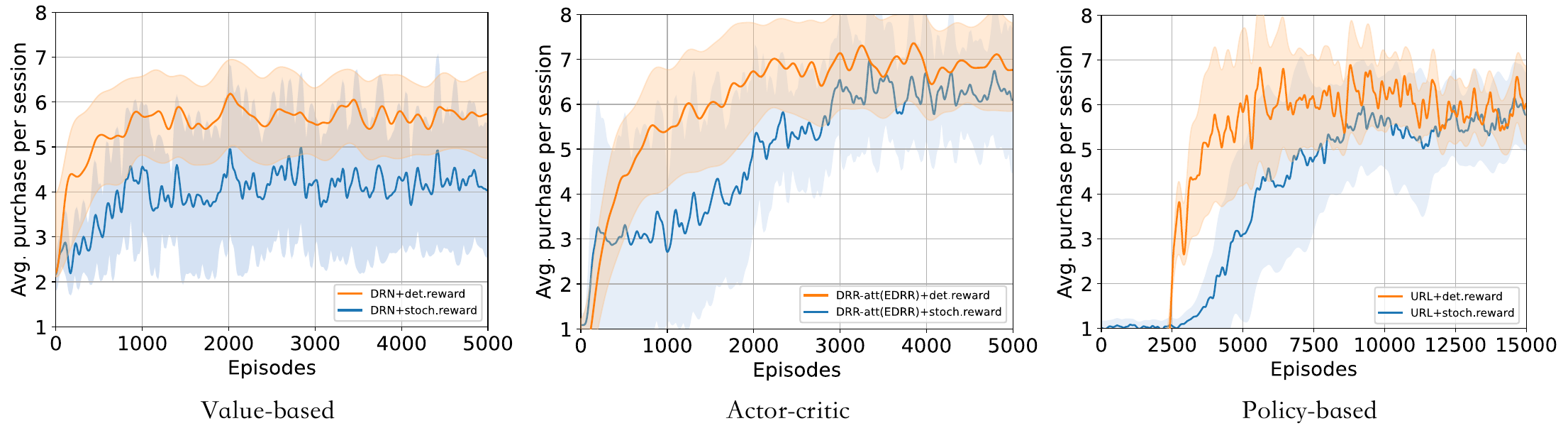}
\end{center}
\vspace{-5mm}
\caption{{Illustration of performance decline due to stochastic rewards on Virtual Taobao.} The plots compare the performance of various model-free RL methods trained with stochastic or deterministic rewards. See experiment details in Section \ref{simulation_experiment}.
\vspace{-2mm}} \label{fig_stochastic_vs_deterministic}
\end{figure*}

Despite the above various model-free methods, most of them do not consider the high stochasticity of rewards, which is a source of sample inefficiency that universally exists in real-world recommendation scenarios. Unlike many classic RL tasks such as Atari 2600 Games \cite{mnih2013playing} and MuJoCo \cite{lillicrap2015continuous} where rewards are deterministic and fully determined by the current state and the selected action, in recommender systems, reward signals of users' behaviors (e.g., click/purchase) are mostly stochastic. Specifically, a user may respond to a given item differently from time to time, relying on unobserved features (such as the surrounding environment) that cannot be captured by the systems. Therefore, in RL-based recommender systems, the high stochasticity of rewards tends to deteriorate the sample efficiency, which is analogous to supervised learning where the stochasticity of signals will slow down the convergence \cite{bottou2018optimization,jeunen2020joint}. 

{In this paper, we first conduct an empirical study in a standard RL recommendation scenario, i.e., the Virtual Taobao \cite{shi2019virtual}, to disclose the detrimental effect of high stochasticity of rewards. As shown in Figure \ref{fig_stochastic_vs_deterministic}, model-free RL methods trained with stochastic rewards suffer from slow convergence and low final performance compared to those trained with deterministic rewards.}

To handle the stochasticity of reward and further improve the sample efficiency for model-free RL-based recommender systems, we propose the Stochastic Reward Stabilization (SRS) framework, which replaces the stochastic reward by its conditional expectation predicted via an extra reward estimation model. Integrating SRS with auxiliary training \cite{liu2020end,chen2021user} results in the SRS with Shared Representation (SRS2) framework, which further accelerates the training of user and item representations. Both frameworks are model-agnostic, as they allow to use any supervised recommendation methods to enhance model-free RL-based recommender systems, which is attractive to industrial practitioners. Extensive experiments on a publicly available recommendation simulator and a real-world billion-user recommendation task show the superiority of the proposed method on sample efficiency and final performance. 

\vspace{-1ex}  

\section{Related work}

\noindent\textbf{Model-free RL.} Most currently used RL-based recommender systems are based on model-free methods, which can be categorized into three classes: value-based methods \cite{zheng2018drn,zhao2021dear}, policy-based methods \cite{chen2019top}, and actor-critic methods \cite{zhao2018deep,chen2022off}. See \cite{chen2021survey} for a detailed survey. As a model-agnostic approach, our proposed SRS can be combined with these methods by replacing instantaneous rewards in Bellman equation (in valued-based/actor-critic methods) or policy gradient calculation (in policy-based methods) with estimated rewards.

\noindent\textbf{Model-based RL.} These methods directly model the environment dynamics \cite{bai2019model,chen2019generative,zou2020pseudo}, which improves sample efficiency. However, it is very difficult to estimate the state transition in many real-world recommendation tasks with large state and action spaces \cite{afsar2021reinforcement}.

\noindent\textbf{RL with Auxiliary Training.} Another approach to improve sample efficiency is auxiliary training, which accelerates the learning of representations via extra tasks. Exemplar methods include DQN-att (EDRR)\citep{liu2020end}, DRR-att (EDRR)\citep{liu2020end}, and URL \citep{chen2021user}. A relevant approach is self-supervised or imitation learning, which also introduces auxiliary tasks to learn better representations \citep{xin2020self,yuan2021improving}. Although we also have a reward estimation task, it is intrinsically different as our task aims to generate a deterministic reward for RL model rather than learning better representations. Note that our method can be combined with auxiliary training, as discussed in Section 4.2.


\section{Problem Formulation}
In RL-based recommender systems, interactions between the user and the recommender system are modeled by Markov Decision Processes (MDPs) \cite{sutton2018reinforcement}. Formally, an MDP can be represented by a tuple $(\sS, \sA, \probP, R, \gamma, \rho_0)$. At each step $t=1,2,...$, the recommender system observes user/context features as the current state $s_t\in \sS$ and chooses an action $a_t\in\sA$ that represents the recommended item according to a policy $\pi_\theta(\cdot|s_t)$ parameterized by $\theta$. 
The user then replies to the system with certain feedback (e.g., click, rating, purchase) taken as the reward $r_{t}$ with probability $ R(r_{t} |s_t, a_t)$, and the state at the next step transfers to $s_{t+1}$ with probability $ \probP(s_{t+1} | s_t,a_t)$. The process goes on until reaching certain terminal states or a pre-defined maximum number of steps $T$. 

The objective of the agent is to find a policy $\pi_\theta$ that maximizes the expected discounted cumulative reward over all users, i.e.,
\begin{align} \label{problem_formulation}
    \max_{\pi_\theta} \E_{\pi_\theta}\left[\sum_{t=0}^\infty \gamma^t r_t  \middle | s_0\sim \rho_0 \right],
\end{align}
where $\rho_0$ is the initial state distribution and $\gamma$ is the discount factor for future reward; the expectation is taken over the above stochastic process, i.e., $a_t\sim\pi_\theta(\cdot|s_t), r_{t}\sim R(\cdot|s_t, a_t), s_{t+1} \sim \probP(\cdot|s_t, a_t)$.

\section{Our Method}

Recall that, as shown in Figure \ref{fig_stochastic_vs_deterministic}, the stochasticity of rewards in recommendation tasks severely degrades the sample efficiency and performance of existing model-free RL methods. In this section, we propose a \underline{\textbf{S}}tochastic \underline{\textbf{R}}eward \underline{\textbf{S}}tabilization (\textbf{SRS}) framework, which stabilizes the reward by replacing the observed reward with its expectation conditioned on the state and action $\hat{r}:= \E(r|s,a)$, where the expectation is estimated via a reward estimation model.

\subsection{Stochastic Reward Stabilization (SRS)} 

\begin{figure}[t] 
\begin{center}
\includegraphics[width=0.475\textwidth]{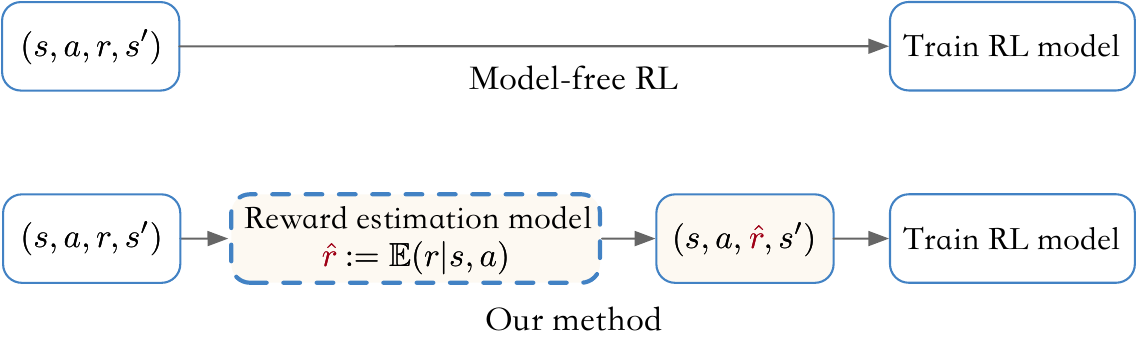}
\end{center}
\vspace{-3mm}
\caption{
Comparison of model-free RL and SRS framework. 
} \label{fig_architecture}
\vspace{-1mm}
\end{figure}

In many recommendation tasks, the immediate reward is usually the click, purchase, or user retention. Many supervised reward estimation techniques has been proposed in recommender systems \cite{koren2009matrix,mcmahan2013ad,rendle2012factorization,huang2013learning,hidasi2015session}. We consider using them to stabilize the reward and propose the SRS framework, which is depicted in Figure \ref{fig_architecture}. With samples $\sD:= \{(s,a,r,s')\}$ containing state $s$, action $a$, immediate reward $r$ and next state $s'$, model-free deep RL methods use the samples directly to learn an optimal policy $\pi_{\theta}(\cdot | s)$. To tackle the sample inefficiency caused by the stochastic reward, we propose a reward stabilization process that replaces the stochastic reward by its conditional expectation $\hat{r}:= \E(r|s,a)$, as illustrated in the dashed block in Figure \ref{fig_architecture}. Then the reconstructed dataset $\hat{\sD}=\{(s,a,\hat{r},s')\}$ is passed to the RL training algorithms. Note that our method only uses a reward estimation model, which is different from model-based methods that require estimating the state transition probability.

The proposed method is model-agnostic, in the sense that \emph{any} supervised recommendation technique can be used to boost the sample efficiency of \emph{any} RL-based recommender system. We here give an example on the classic DQN method. Recall that value-based methods evaluate the Q value function, i.e., the long-term expected reward of taking action $a$ at state $s$ and following policy $\pi_\theta$. Replacing $r$ with $\hat r$ in its form, the modified Q function becomes
\begin{align}
\hat Q_{\pi_\theta}(s,a) := \E_{\pi_\theta}\left[\sum_{t=0}^\infty \gamma^t \hat r_t \middle | s_0 = s, a_0 = a\right].
\end{align}
Then the agent selects the action to maximize the modified Q function at each step, i.e., $a\in\argmax_a \hat Q_{\pi_\theta}(s,a)$. In vanilla DQN, the Q function is learned from the collected dataset $\sD$ via the temporal difference approach \cite{sutton2018reinforcement}. Applying SRS, the loss function becomes 
\begin{align} \label{eqt_dqn}
  L_{DQN} := \E_{(s,a,\hat r,s')\sim\hat\sD} [(y - \hat Q_{\pi_{\theta}}(s, a))^2],
\end{align}
where $y = \hat r + \argmax_{a'} \gamma \hat Q_{\pi_{\theta'}}(s', a')$ is the estimated modified Q value predicted by the target network with parameters $\theta'$ \cite{mnih2013playing}.  

Besides valued-based RL, other branches of model-free RL, namely actor-critic and policy-based methods, can also be combined with SRS by optimizing the strategy $\pi_\theta$ based on the modified dataset $\hat\sD$ instead of the collected dataset $\sD$. Their detailed formulations are omitted from this paper due to the space limit.

\begin{figure*}[tb]
\begin{minipage}[c]{0.25\textwidth}
\includegraphics[width=\textwidth]{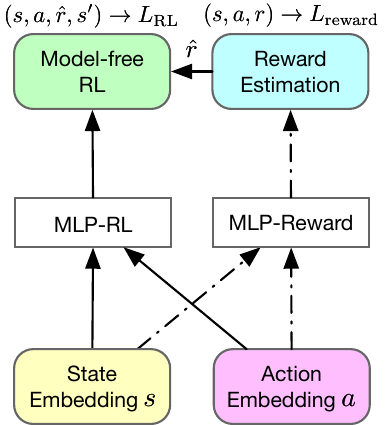}
\caption{Network structure of the SRS2 method. In SRS2, the state embedding and action embedding are shared, which are trained using the gradients of the reward estimation model.}
\label{fig_model_simple}
\end{minipage}
\hfill
\begin{minipage}[c]{0.73\textwidth}
\includegraphics[width=\textwidth]{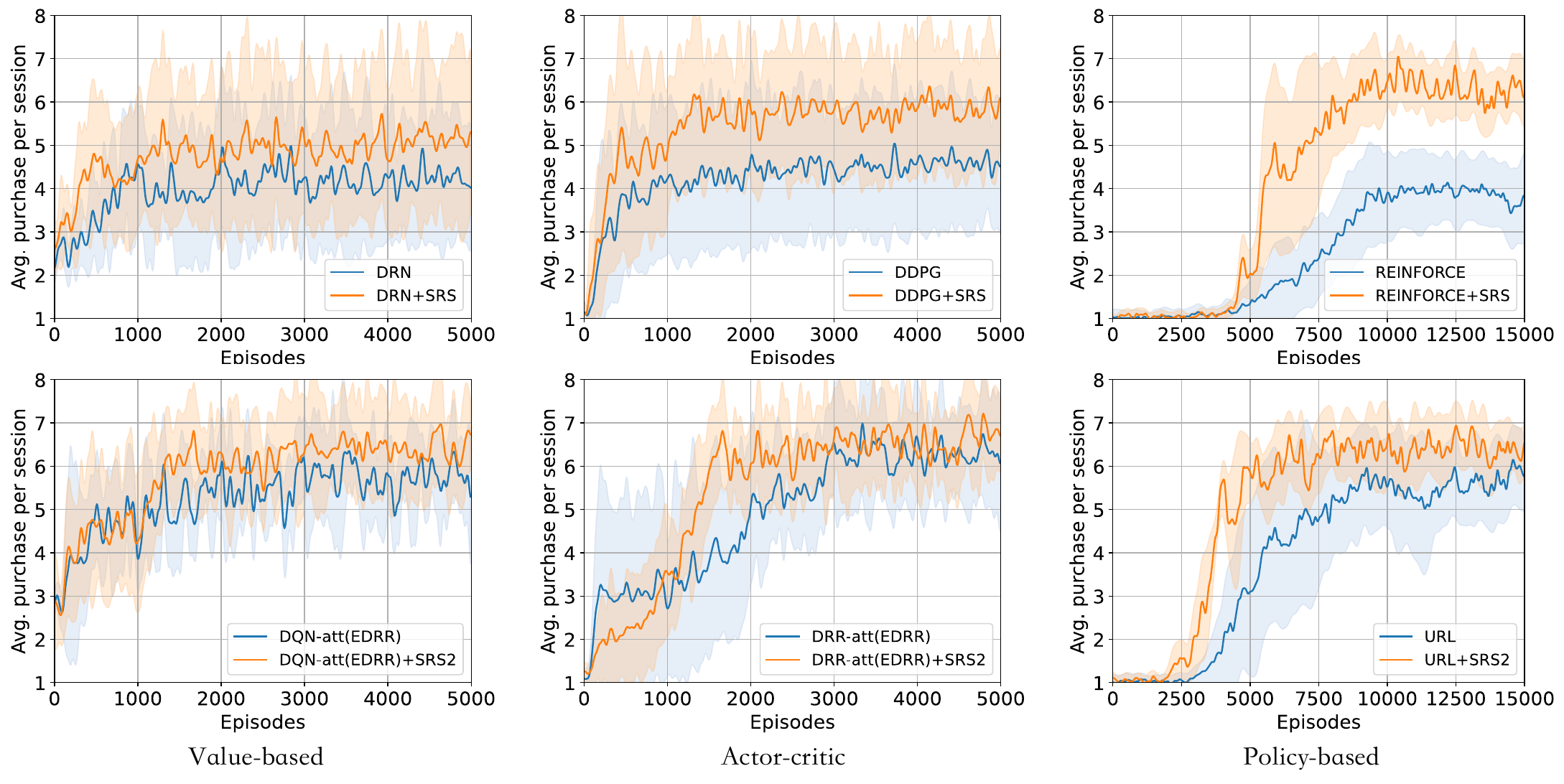}
\vspace{-8.5mm}
\caption{Performance comparison of various model-free RL methods \emph{with/without} reward stabilization on Virtual Taobao. The two rows of subplots present the performance of the three most commonly used RL methods and three SOTA models, respectively.}
\label{fig_sample_efficiency}
\end{minipage}
\end{figure*}

\subsection{SRS with Shared Representation (SRS2)}

As mentioned before, current model-free RL-based recommender systems often share the representation of states and actions of the RL task with an auxiliary task, which substantially accelerates the training of the representations, especially in scenarios with large state and action spaces \cite{cheng2016wide,liu2018deep,zhao2020whole,zhao2018deep,zheng2018drn}. Here we incorporate this idea into SRS, which results in the \underline{\textbf{SRS}} with \underline{\textbf{S}}hared \underline{\textbf{R}}epresentation (\textbf{SRS2}) framework, achieving the merits of both worlds.

As shown in Figure \ref{fig_model_simple}, the reward estimation and the RL models share a common embedding, which maps each user or item features to a state embedding $s$ or an action embedding $a$, respectively. The embedding vectors $s$ and $a$ are concatenated and fed into the subsequent RL and reward estimation modules. At the training stage, since the supervised signal of the reward estimation task is more stable than the long-term signal of the RL task \cite{liu2020end}, we only use gradients of the supervised task to update the embedding layer. 

Note that from the perspective of representation learning, our introduced reward estimation task can be understood as a kind of auxiliary task, which is previously used to accelerate the learning of embeddings \cite{liu2020end,chen2021user}. However, in SRS2, the output signal $\hat r$ of the task is further used to guide the training of the RL module, which is a core difference from previous methods. The double usages of the reward estimation module in SRS2 combine the benefits of SRS and shared representation, which attains better convergence and final performance than previous auxiliary training methods empirically.

\section{Experiments} \label{simulation_experiment} This section evaluates the efficacy of our proposed frameworks.

\subsection{Experiment Setup}

\subsubsection{Experimental Environments.} Our experiments are conducted in a simulation environment and a live environment.

\noindent\textbf{Simulation Environment.}~ We conduct simulation experiments on a publicly available simulator, i.e., Virtual Taobao \cite{shi2019virtual}\footnote{Virtual Taobao source code available at: \href{https://github.com/eyounx/VirtualTaobao}{https://github.com/eyounx/VirtualTaobao}}. In each round, the search engine receives a search request from a user and responds to the user with some items to be displayed. Then the user gives certain feedback. The reward is set to be $1$ if the user purchases any recommended item and $0$ otherwise. Specifically, for an estimated user purchase with probability $r$ ($r\in[0, 1]$), its deterministic reward $r_{det}$ is directly set to be $r$, while its stochastic reward $r_{stoch}$ is randomly sampled from a Bernoulli distribution parameterized by $r$, i.e., $r_{stoch}=1$ \emph{w.p.} $r$ and $r_{stoch}=0$ \emph{w.p.} $1-r$. 

\noindent\textbf{Live Environment.}~ We apply our proposed method to a real-world coupon recommendation task with more than 1 billion participants. The recommender system adopts a two-step scheme commonly used in industrial recommendation scenarios \cite{covington2016deep,ie2019slateq}: (i) a candidate generator retrieves hundreds of coupons from all available items; (ii) a candidate ranking network generates the final coupon recommended to the user. Our method is applied to the second step.

In our experiments, we adopt the commonly used myopic DIN model \cite{zhou2018deep} to estimate the expected immediate reward, which is co-trained with the RL model on the same dataset.

\subsubsection{Compared Methods.} For each class of model-free RL, we consider a classic method and a SOTA method, respectively.

\noindent\textbf{Value-based Approach.}~ We consider \textbf{DRN} \cite{zheng2018drn} that uses double Q learning \cite{van2015deep} and dueling network \cite{wang2016dueling} to tackle overestimation of DQN, and \textbf{DQN-att(EDRR)} \cite{liu2020end} (SOTA) that introduces an auxiliary training task to learn a better user representation. 

\noindent\textbf{Policy-based Approach.}~ We employ \textbf{REINFORCE} \cite{chen2019top,afsar2021reinforcement} as a classic method in this approach, and \textbf{URL} \cite{chen2021user} (SOTA) that learns the state and action representations concurrently via auxiliary training. 

\noindent\textbf{Actor-critic Approach.}~ We examine \textbf{DDPG} that is widely used by many recommender systems \cite{zhao2018deep,gao2022value,xiao2021general}, and \textbf{DRR-att(EDRR)} \cite{liu2020end} (SOTA) that uses the DQN-att(EDRR) as the critic. 

\subsection {Experimental Results}

\subsubsection{Simulation Experiment}
We first compare the performance and sample complexity in the simulation environment.

\noindent\textbf{Performance.}~We adopt the average cumulative reward as the performance metric, which measures the number of purchases in each session. For each method, we compare its vanilla version and that combined with SRS/SRS2. We evaluate each model every $10$ steps \cite{zou2020pseudo} and present the results in Figure \ref{fig_sample_efficiency}, from which we have two main observations. First, combining each method with SRS achieves a higher reward. The performance gain universally exists, indicating that SRS is agnostic to the RL model. Second, the average reward of each method with SRS has a lower variance than the vanilla one, especially on actor-critic and policy-based approaches. This accords well with our intuition that implanting a supervised model in RL stabilizes the reward as well as the learning process. 

\noindent\textbf{Sample Efficiency.}~For each method, we measure the number of training samples processed until the performance reaches some certain satisfactory threshold \cite{kaiser2019model}. In this experiment, we set the threshold to be the number of steps at which the reward attains $7$ for the fifth time. Table \ref{table_compare} presents the result and its $95\%$ confidence interval for each method after repeating running for ten times. From the results, we observe that SRS substantially promotes sample efficiency in all RL approaches. In particular, on actor-critic and policy-based approaches, SRS2 achieves at least $2\times$ speedup in training compared to SOTA models. Note that on the DQN-based approach, the acceleration is not that significant compared to the above two approaches, possibly because the value-based approach is by itself more robust to the stochastic noise than other approaches.

\begin{table}[t] \label{table_compare}
\vspace{2mm}
  \caption{Sample efficiency comparison of various methods \emph{with/without} reward stabilization. We measure the number (in thousands) of episodes to reach a certain performance threshold. Note that vanilla DRN, DDPG, and REINFORCE never attain the threshold, indicating very low efficiency.}
  \label{table_compare}
  \begin{center}
  \begin{tabular}{ccc} 
  \toprule 
  Type & Method & Sample Efficiency \\  \midrule
\multirow{4}{*}{Value-based}
    & DRN   & - \\
    & DRN + SRS  & 3.30 ($\pm$ 1.66) \\   \cline{2-3}
    & DQN-att(EDRR) & 2.07 ($\pm$ 0.34) \\
    & DQN-att(EDRR) + SRS2 & \textbf{1.62 ($\pm$ 0.18)} \\ \midrule

\multirow{4}{*}{Actor-critic}
    & DDPG  & - \\
    & DDPG + SRS & 2.94 ($\pm$ 0.52) \\  \cline{2-3}
    & DRR-att(EDRR) & 3.28 ($\pm$ 0.20) \\
    & DRR-att(EDRR) + SRS2 & \textbf{1.67 ($\pm$ 0.09)} \\ \midrule
\multirow{4}{*}{Policy-based}
    & REINFORCE & - \\
    & REINFORCE + SRS & 7.56 ($\pm$ 0.47) \\  \cline{2-3}
    & URL & 9.43 ($\pm$ 0.44) \\
    & URL + SRS2 & \textbf{4.87 ($\pm$ 0.33)} \\
    \bottomrule
\end{tabular}
\end{center}
\end{table}

\subsubsection{Live Experiment} We adopt two performance metrics commonly used in industrial scenarios, i.e., the user return and the user payment. The former metric measures the average number of times the user comes back to the coupon recommendation scenario during the whole experimental period, which reflects the user's long-term satisfaction. The latter metric measures the average number of payments with coupon redemption for each user. 

We compare the performance of DQN-att(EDRR) equipped with SRS2 and the myopic DIN method via online A/B testing. The two-week live traffic with more than 10 million participants shows that our method achieves a $+3.16\% (\pm0.26\%)$ relative improvement on user return and a $+0.89\% (\pm 0.17\%)$ improvement on user payments, showing that our proposed method is very effective in industrial-level real-world recommendation tasks.

\section{Conclusions}
In this paper, we first discover the detrimental effect of stochastic rewards on RL-based recommender systems via empirical study. To resolve the issue, we design a stochastic reward stabilization framework that replaces the stochastic signal with the signal predicted by a reward estimation model, then combines it with shared representations. Extensive experiments show the superiority of our proposed methods on sample efficiency and final performance.

\clearpage

\bibliographystyle{ACM-Reference-Format}
\balance
\bibliography{samples/sample-base}

\clearpage

\appendix

\end{document}